\documentclass[11pt]{article}
\usepackage{acl2016}
\usepackage{times}
\usepackage{url}
\usepackage{latexsym}
\usepackage{color}
\usepackage{graphicx}
\usepackage{float}

\aclfinalcopy 
\def\aclpaperid{22} 

\newcommand{\snap}{SnapToGrid}

\title{SnapToGrid: From Statistical to Interpretable Models for
  Biomedical Information Extraction}

\author{Marco A. Valenzuela-Esc\'{a}rcega, Gus Hahn-Powell, Dane Bell, Mihai Surdeanu \\
	 University of Arizona \\
         Tucson, AZ 85721, USA \\
         {\tt \{marcov, hahnpowell, dane, msurdeanu\}@email.arizona.edu}} 

\date{}

\begin{document}
\maketitle
\begin{abstract}

We propose an approach for biomedical information extraction that marries the advantages of machine learning models, e.g., learning directly from data, with the benefits of rule-based approaches, e.g., interpretability. Our approach starts by training a feature-based statistical model, then converts this model to a rule-based variant by converting its features to rules, and ``snapping to grid'' the feature weights to discrete votes. In doing so, our proposal takes advantage of the large body of work in machine learning, but it produces an interpretable model, which can be directly edited by experts. 
We evaluate our approach on the BioNLP 2009 event extraction task. Our results show that there is a small performance penalty when converting the statistical model to rules, but the gain in interpretability compensates for that: with minimal effort, human experts improve this model to have similar performance to the statistical model that served as starting point.

\end{abstract}

\section{Introduction}

Due to the deluge of unstructured data, information extraction (IE)
systems, which aim to translate this data to structured information,
have become ubiquitous. For example, applications of IE range from
parsing literature~\cite{iyyer2016} to converting thousands of cancer
research publications into complex proteins signaling
pathways~\cite{cohen2015}.

By and large, in academia most of these approaches are implemented
using machine learning (ML). This choice is warranted: generally, ML
approaches, where the machine learns directly from the data, perform
better than approaches where human domain experts encode the structure
to be extracted manually. For example, the top systems in the BioNLP
event extraction shared tasks have consistently been ML-based
approaches~\cite{bionlp2009,bionlp2013}.
However, this is only part of the story: most of these models cannot
be easily understood by their users, and, by and large, cannot be
modified without retraining. This ``technical debt'' of
ML~\cite{sculley2014} is better understood in industry: Chiticariu et
al.~\shortcite{chiticariu2013} report that 67\% of large commercial
vendors of natural language processing (NLP) software focus on
rule-based IE, and an additional 17\% on hybrid systems that combine
rule-based and ML approaches.

In this paper we focus on {\em interpretable models} for information
extraction, i.e., models that: (a) can be understood by human users,
and (b) can be directly edited and improved by these users. In
particular, we focus on deterministic, rule-based models. Here, we
introduce a novel approach to generate such models, which maintains
both the advantages of ML such as learning from data, and the benefits
of interpretability such as allowing human domain experts to directly
edit and improve these models. Specifically, our contributions are:
{\flushleft {\bf (1)}} We introduce a simple strategy that converts statistical models
  for IE to rule-based models. We call the proposed algorithm
  \snap. Our approach works in three steps. First, we train a
  statistical model for the task at hand. Here we experiment with logistic regression,
  but the proposed method is, in principle, independent of the underlying statistical model.
  Further, our strategy can
  operate over multiple classifiers that are part of the same IE
  system (e.g., one classifier to identify event triggers, and another to identify event arguments). 
  Second, we convert features to rules implemented in Odin, a modern
  declarative rule language~\cite{valenzuela2016,valenzuela2015}. We
  also discard most of the statistical information acquired
  previously, by converting feature weights to discrete votes, which
  guarantees interpretability (hence the SnapToGrid name). Third,
  human domain experts inspect and manually improve the generated
  model, under certain time constraints.

{\flushleft {\bf (1)}} We evaluate our approach on the BioNLP 2009 core event
  extraction task, and demonstrate that the resulting interpretable
  model has similar performance to the statistical model that served
  as starting point. 

\section{Approach}

Our motivation for this work is to keep the human domain expert in the
loop when building IE systems. We show in Section~\ref{sec:results}
that this is beneficial, even when the domain experts have limited
time to work on the task and no access to data other than the model
itself. To achieve this ``human in the loop'' goal we propose the
following three-step algorithm:

\begin{figure*}[t]
\centering 
\includegraphics[width=0.66\textwidth]{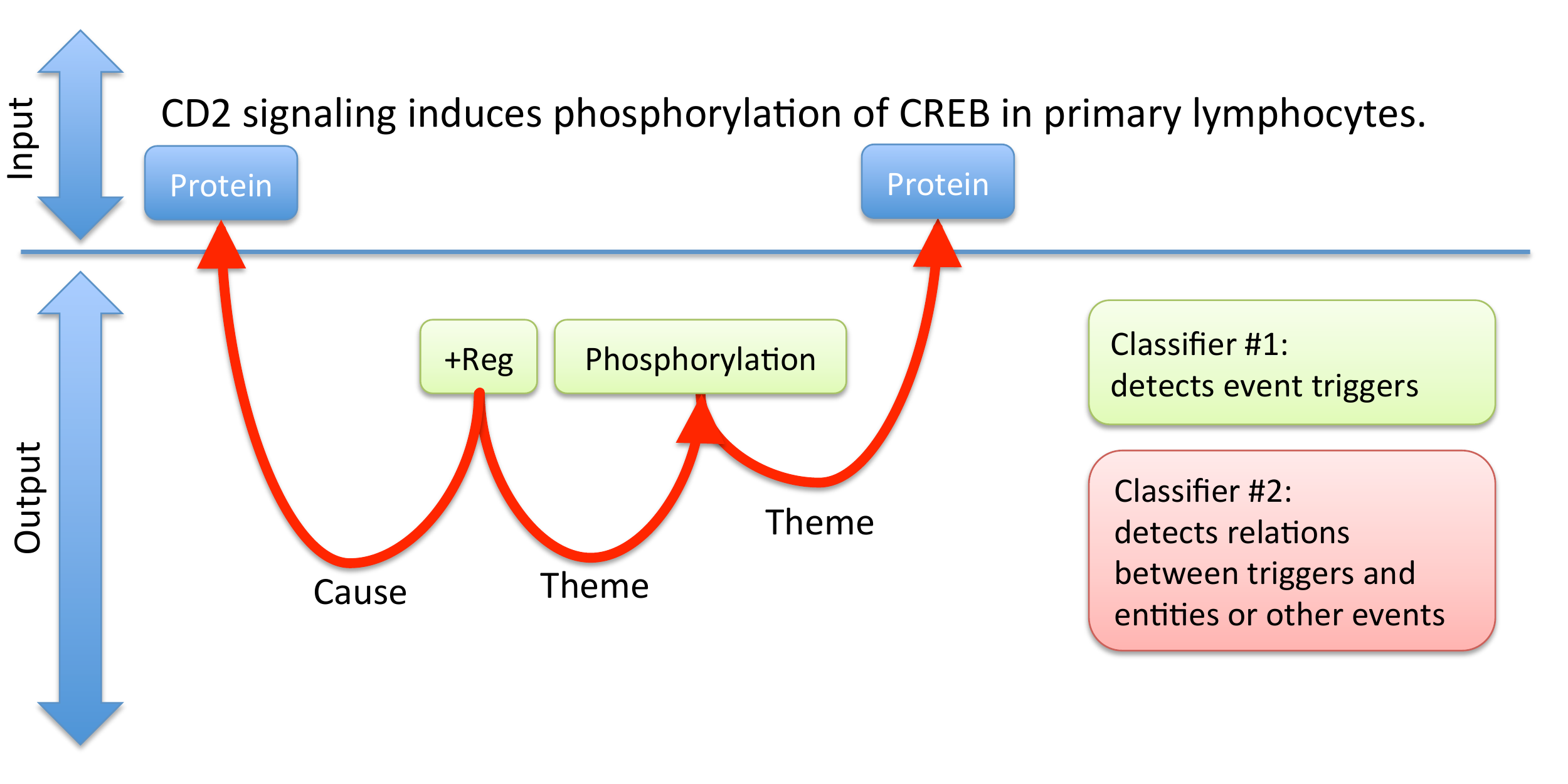}
\vspace{-2mm}
\caption{Architecture of the statistical model for the BioNLP core
  event extraction task.}
\vspace{-4mm}
\label{fig:arch}
\end{figure*}

\begin{enumerate}
\item Train a statistical model for the IE task at hand
  (Section~\ref{sec:statmodel}). The model may consist of several
  statistical classifiers. For example, for the BioNLP event
  extraction task, the most common approach involves two classifiers:
  one to identify event triggers, and a following classifier to
  identify event participants. One restriction is that these
  classifiers be feature-based classifiers, e.g., logistic regression,
  rather than the classifiers based on latent representations, e.g.,
  neural networks.
\item Convert the statistical model into an interpretable, rule-based
  model (Section~\ref{sec:convert}):
\begin{enumerate}
\item First, we convert the features to rules in the Odin language.
\item Then, we assign to each rules ``votes'' for a given class, by
  ``snapping to grid'', i.e., converting to discrete values, the
  weights computed by the above statistical model.
\end{enumerate}
\item Domain experts edit the produced rule-based model directly,
  aiming to improve its quality with respect to both coverage and
  precision (Section~\ref{sec:editmodel}).
\end{enumerate}

We detail this process in the rest of this section, focusing on the
BioNLP core event extraction task as the domain of interest.

\subsection{Step 1: Build Statistical Model}
\label{sec:statmodel}

Our statistical model is inspired by the top performing approach at
the 2009 evaluation~\cite{bjorne2009extracting}. The approach is
summarized in Figure~\ref{fig:arch}. Similar
to~\cite{bjorne2009extracting}, our approach consists of two
classifiers: the first classifier detects and labels event trigger
words in the input text; the second classifiers extracts and labels
relations between event triggers and potential event participants,
which can be either Protein entities or other event triggers. Both
classifiers are implemented using multi-class logistic regression
(LR), but our conversion process (Steps 2 and 3) 
is independent of the underlying statistical model, so, in principle, other 
feature-based classifiers that assign
explicit weights to features could be used, e.g., perceptron,
or linear support vector machines. 

\subsubsection*{The Trigger Classifier}

The first classifier sequentially labels each word in the input text
as a trigger for a specific BioNLP event class, or as {\tt Nil}
otherwise. We implemented the following features:

\begin{description}
\item[Surface features:] These features include the original and
  lemmatized words, and the presence of the word in a gazetteer of
  known event triggers (constructed automatically from the training
  data). These features are generated for the word being classified,
  as well as the words surrounding it inside a window of $n$
  tokens. 
  We used two windows in our experiments, with $n=1$ and $n=4$.
  Further, bag-of-words features 
  are generated for the windows and for the sentence as a whole.
\item[Syntactic features:] These features capture the syntactic dependencies (both incoming and outgoing) directly
  connected to the token. All syntactic information was represented using
  Stanford dependencies~\cite{de2008stanford}, and was generated using
  the CoreNLP toolkit~\cite{manning2014stanford}.  For each of these
  paths, we generate two different versions: one containing just the
  label and direction of the syntactic dependencies, and another
  including also the destination words. 
\item[Entity features:] These features encode the number of other
  entities surrounding the token, both inside a window and in the
  sentence as a whole.
\end{description}


\subsubsection*{The Event Participant Classifier}

This classifier pairs all the triggers detected by the previous
classifier with other named entities (Proteins in this case) or event triggers that occur in the
same sentence. These pairs are then classified into one of the
possible participant relations, or {\tt Nil} indicating that there is
no relation between the pair.  This classifier uses the following
features:

\begin{description}
\item[Syntactic features:] These features are based on the shortest
  path connecting the two mentions (trigger and candidate participant)
  in the Stanford syntactic dependency graph. Two versions of the
  shortest path are used: a lexicalized one (capturing the words along
  the path), and an unlexicalized one.
  
\item[Surface features:] These features include: the order of the two
  mentions in text, their distance in terms of tokens, the number of
  entities and triggers in the sentence, the parts of speech and words
  of the mentions, and the number of triggers and entities between the
  mentions.
\item[Consistency features:] These features encode the labels of the
  two mentions jointly, as well as the labels of their
  superclasses. For example, the features $<$Regulation,
  Phosphorylation$>$ and $<$Regulation, Event$>$ are generated for a
  relation between a Regulation event trigger and a Phosphorylation trigger as its theme. These feature
  capture selectional preferences for arguments, e.g., the Theme of a
  regulation event should be another event.
\item[Graph features:] The parent, children, and siblings of the
  mentions in the syntactic dependency graph.
\end{description}

\subsubsection*{Limitations}
\label{sec:limits}

Not all of the above features can be represented as rules in the
current implementation of the chosen rule language.
Currently\footnote{As of June 2016}, Odin rules capture paths (over sequences or directed graphs) that are anchored at both ends (e.g., from an event trigger to an event argument)~\cite{valenzuela2015,valenzuela2016}.  
  Because of this, Odin cannot represent the
following information: bag-of-word features, syntactic paths that are
not anchored at both ends (such as dependencies connected only to
event trigger candidates), and features that
count occurrences of tokens or entities in text.
In Section~\ref{sec:results} we analyze the performance drop when such
features are removed from the model.

\subsection{Step 2: Convert the Statistical Model to a Rule-based Model}
\label{sec:convert}

Once the statistical model is constructed, we employ the lossy process
below to convert it to an interpretable one.

\subsubsection*{Converting Features to Rules}
\label{sec:convertrules}

\begin{figure}[t]
  \centering
  \includegraphics[width=0.66\columnwidth]{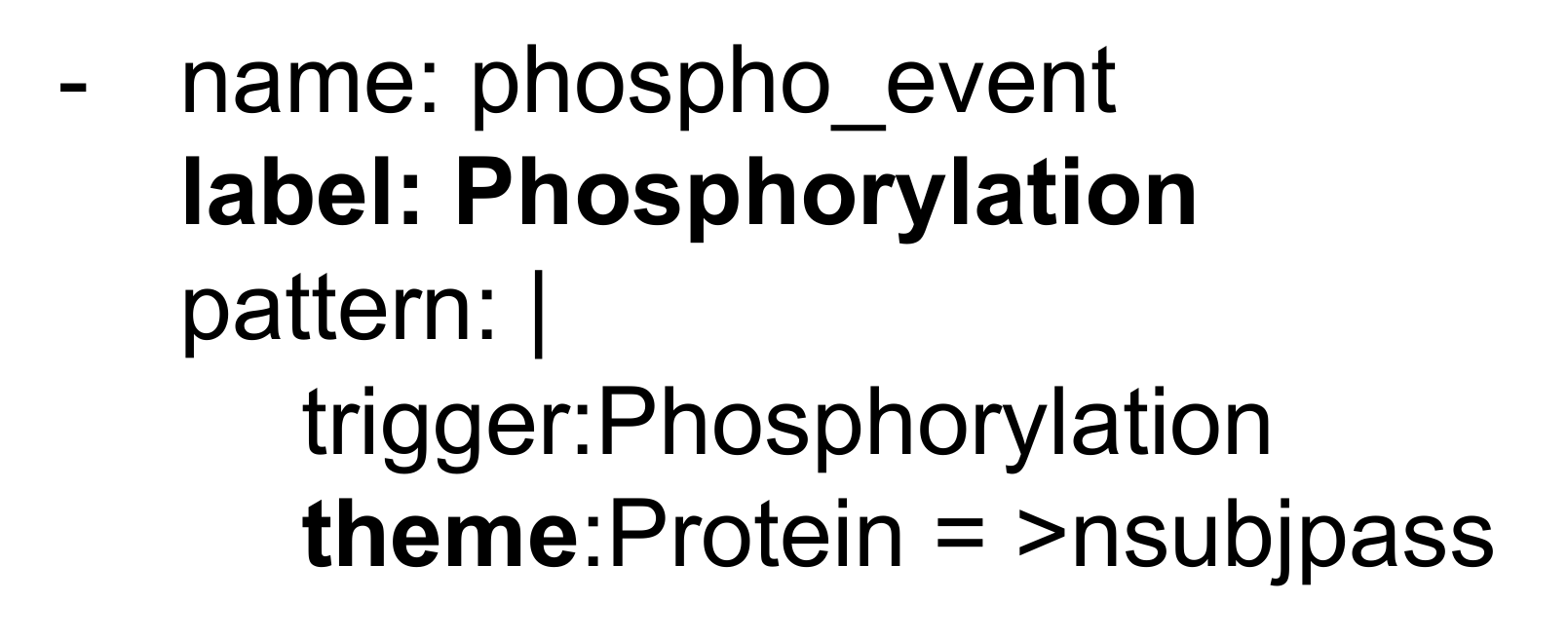}
  \vspace{-2mm}
  \caption{Example of a rule for event participant classification that
    is built from a single feature. The feature captures the passive
    nominal subject ({\tt nsubjpass}) outgoing ({\tt >}) from a
    Phosphorylation trigger and landing on a Protein. The bold font
    indicates the rule output, i.e., the nominal subject is the {\bf
      theme} of a {\bf Phosphorylation} event. }
  \label{fig:rulebuilding}
  \vspace{-4mm}
\end{figure}

First, we convert the features encoded in the statistical model to
rules in the Odin language~\cite{valenzuela2015,valenzuela2016}. In general, the features previously
introduced consist of conjunctions of information bits, each of which
corresponds to a different rule fragment. For example, for the
classification of event participants, one such conjunction captures
the type of the expected trigger (e.g., Phosphorylation), combined
with the syntactic path that connects the trigger with the participant candidate (e.g., an
outgoing passive nominal subject -- {\tt nsubjpass}), and a semantic
constraint for the type of named entity of the participant (e.g.,
Protein). These are immediately translatable to Odin rules, as
illustrated in Figure~\ref{fig:rulebuilding}.

Importantly, the rules encode output information as well, e.g., the
recognized event participant serves as a theme for a Phosphorylation
event in Figure~\ref{fig:rulebuilding}. At this stage, this
information is exhaustively generated from all possible classifier
labels (e.g., for the classification of event participants these
labels are the cartesian product of \{{\tt theme}, {\tt cause}\} and
possible event labels \{{\tt Phosphorylation}, {\tt Binding},
\dots\}). Of course, some of these outputs do not apply. For example,
it is highly unlikely that the rule shown in
Figure~\ref{fig:rulebuilding} produces the cause of a Regulation
event.  We quantify the confidence in these outputs in the next stage
of the algorithm.

\subsubsection*{Converting Weights to Votes}
\label{sec:convertweights}

\begin{figure}[t]
  \centering
  \begin{tabular}{c}
  \includegraphics[width=0.95\columnwidth]{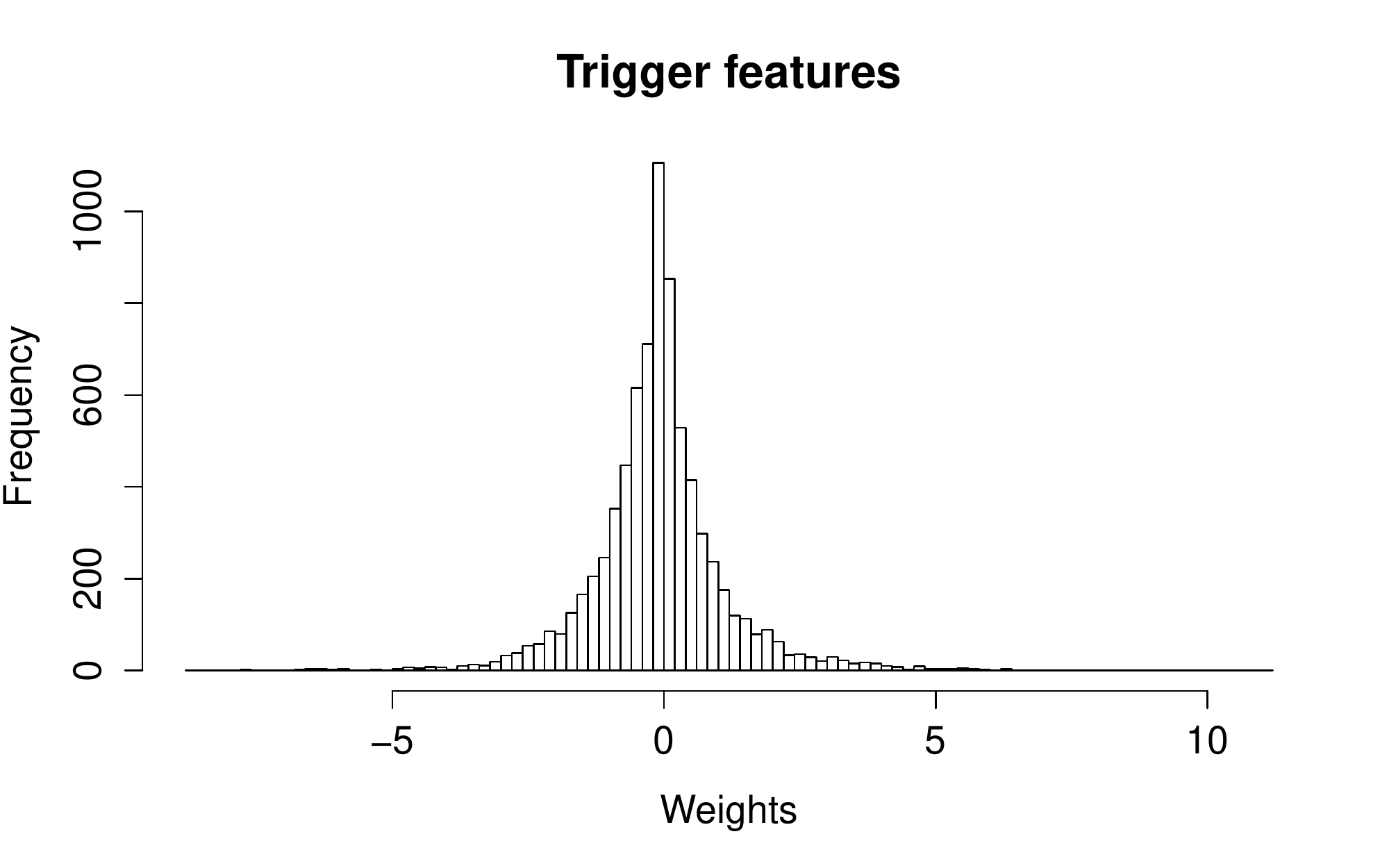} \\
  \includegraphics[width=0.95\columnwidth]{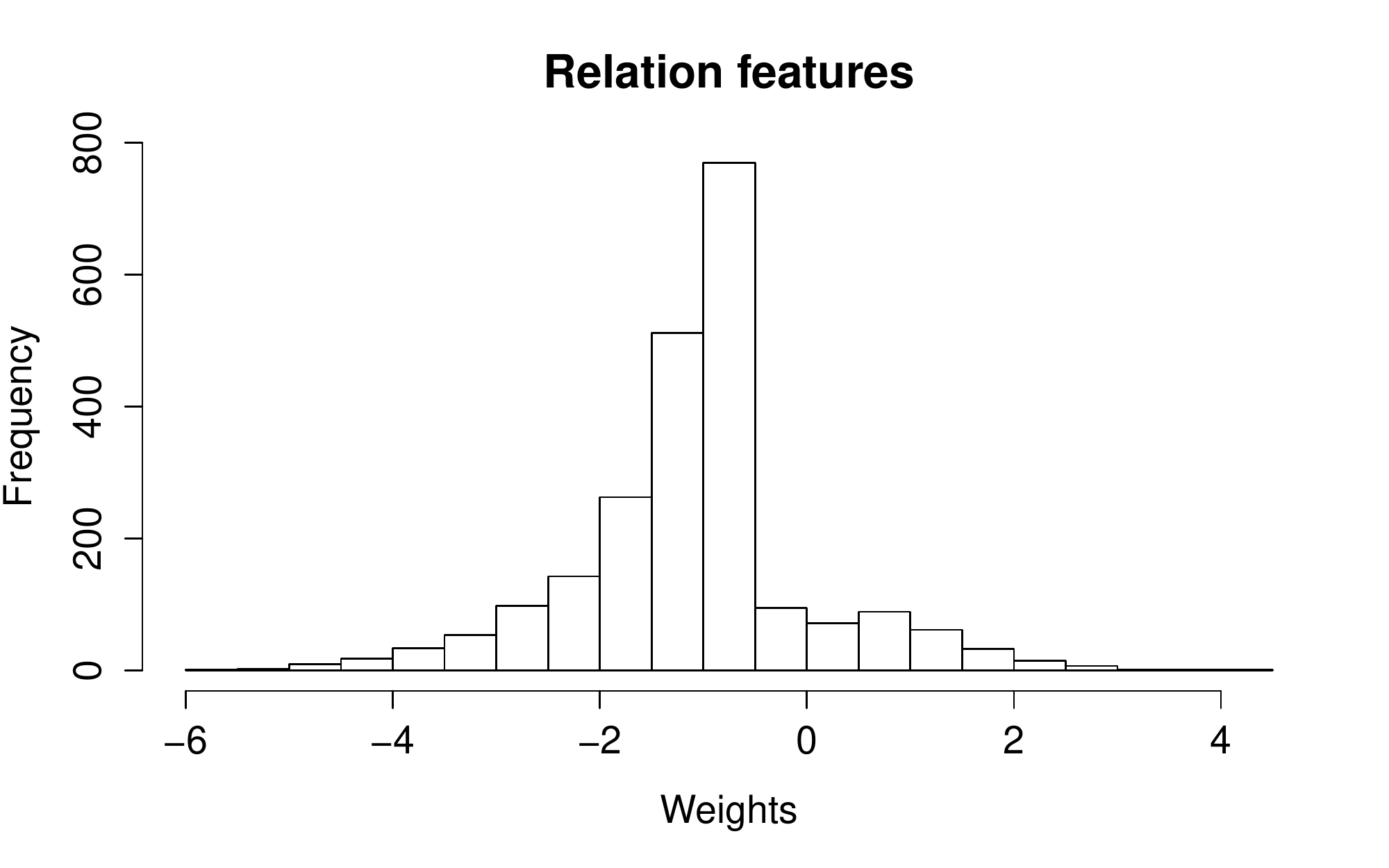}  
  \end{tabular}  
  \caption{Weights of the two classifiers converted to votes (trigger classifier --
    top, participant classifier -- bottom). Each histogram bin receives a number of
    votes (positive or negative) equal to its offset from 0.}
  \label{fig:hist}
  \vspace{-4mm}
\end{figure}

Feature weights are unbounded continuous values that are difficult to
interpret and manually modify. For this reason, we would ideally
prefer to exclude them completely from the interpretable
model. Conceptually, this is simple: we could use the weights to
choose the most likely output label for a rule (from the options
generated previously), and discard them afterwards.  However, our
early experiments demonstrated that this performs poorly, because it
forces the algorithm to ignore the inherent ambiguity of language,
which is captured by the statistical model through weights.
For example, the trigger classifier learns that ``recruits'' serves
as trigger for two different events, Binding and Localization, and,
consequently, assigns different weights to the two labels based on the
amount of evidence seen in training. During inference, the most likely
class is chosen by aggregating the weights of all features that apply.

Given this observation, we chose to preserve the weights, but convert them from the original unbounded continuous values to discrete ``votes'' (positive or negative) that are then used during inference to resolve conflicts. 
This achieves two things. First, we increase the interpretability of the model: humans can now interpret these discrete votes, which mimic a Likert scale~\cite{likert1932technique}. Second, by keeping and using these discretized votes, we preserve some of the statistical power of the model. We show in Section~\ref{sec:results} that some performance is indeed lost in this conversion, but the loss is small and the gain in interpretability compensates for that.

The conversion from continuous weights to discrete votes is a process similar to choosing the bins in a histogram. In our case, we first construct a histogram of all feature weights. Then, each histogram bin receives a number of votes equal to its offset (positive or negative) from 0. For example, all the weights in the second bin to the left of 0 receive two negative votes.
Several methods have been
proposed for selecting the number of bins in a histogram, for
example~\cite{sturges1926choice,doane1976aesthetic,freedman1981histogram}. 
Here, we
use the formula proposed by~\cite{scott1979optimal}:
\begin{equation}
  h = 3.5 \hat{\sigma} n^{-1/3}
\end{equation}
where $h$ is the estimated bin width, $n$ is the sample size, and
$\hat{\sigma}$ is the estimated standard deviation. We chose this formula
because it gives a good compromise
between retaining most of the information in the weights while
minimizing the number of bins. 
The resulting binned weights for trigger and
relation features (generated using the BioNLP 2009 training corpus) are shown in
Figure~\ref{fig:hist}.

\subsection{Step 3: Edit the Rule-based Model }
\label{sec:editmodel}

The output of the previous two steps is a model consisting of a set of rules. The association between rules and output classes is measured  through votes that each matching rule gives to each output label.  The last step in our proposed approach is to let human domain experts improve this model by directly editing it. The experts had complete freedom in the operations they were allowed to do. For example, they could improve the syntactic paths captured by the rules, or increase/decrease the number of votes assigned to a specific rule. The only constraints were: (a) they were not allowed to look only at the learned rules and not at the training data, and (b) they had to complete the process within one hour. This setting is of course artificial and unrealistic. We enforced it in this work to demonstrate the interpretability of the generated model.  

\section{Empirical Results}

We analyze the performance of our approach on the core event extraction dataset from the BioNLP 2009 shared task~\cite{bionlp2009}. All the results reported in this section were measured on the {\em development} partition of the dataset, which was not used at all during training.\footnote{The online scoring website, which would have allowed us to also obtain scores on the official test partition, was down due to updates during the development of this work.} To minimize overfitting, we did not implement any feature selection or other hyper parameter tuning process.

\label{sec:results}

\begin{table}[t]
  \centering
  \begin{small}
  \begin{tabular}{l|r|r|r}
    Event Class & Recall & Precision & F1 \\
    \hline
    Gene\_expression & 67.70 & 68.08 & 67.89 \\
    Transcription & 57.32 & 50.00 & 53.41 \\
    Protein\_catabolism & 71.43 & 68.18 & 69.77 \\
    Phosphorylation & 68.09 & 68.09 & 68.09 \\
    Localization & 69.81 & 74.00 & 71.84 \\
    Binding & 31.85 & 25.57 & 28.37 \\
    \hline
    Event Total & 55.89 & 51.48 & 53.59 \\
    \hline
    Regulation & 17.16 & 33.33 & 22.66 \\
    Positive\_regulation & 19.45 & 41.67 & 26.52 \\
    Negative\_regulation & 14.29 & 36.36 & 20.51 \\
    \hline
    Regulation Total & 18.02 & 39.16 & 24.69 \\
    \hline
    All Total & 35.10 & 47.29 & 40.30 \\
  \end{tabular}
  \end{small}
  \caption{Performance of the statistical model using $L_2$-regularized LR, and all available features.}
  \label{tbl:res_lrl2_all_feats}
\end{table}

Table~\ref{tbl:res_lrl2} lists the results of the complete statistical model, i.e., using all features introduced in Section~\ref{sec:statmodel}, trained using $L_2$-regularized LR. This configuration generated 1,190,029 features with non-zero weights. The table shows that this model achieved an overall F1 score of over 40 points, which likely puts it in the top 5 or 6 (out of 24) systems that participated in the actual challenge.\footnote{\cite{bionlp2009} report results on the official test partition, which are not directly comparable with our results. However, in the authors' experience, the difference in scores between the development and test partitions in this dataset tend to be small. Since the 2009 evaluation, several works have improved upon these results, with performance reaching 58 F1 points, but using more complex methods, including joint inference, coreference resolution, and domain adaptation~\cite{miwa2012boosting,bui2012robust,venugopal2014relieving}.} The performance of this system could be further improved by adding more features proposed in other event extraction approaches~\cite{miwa2010event}, feature selection, hyper parameter tuning, etc.

\begin{table}[t]
  \centering
  \begin{small}
  \begin{tabular}{l|r|r|r}
    Event Class & Recall & Precision & F1 \\
    \hline
    Gene\_expression & 57.58 & 74.28 & 64.87 \\
    Transcription & 40.24 & 57.89 & 47.48 \\
    Protein\_catabolism & 61.90 & 86.67 & 72.22 \\
    Phosphorylation & 51.06 & 82.76 & 63.16 \\
    Localization & 47.17 & 92.59 & 62.50 \\
    Binding & 18.15 & 34.62 & 23.81 \\
    \hline
    Event Total & 42.75 & 64.61 & 51.45 \\
    \hline
    Regulation & 8.28 & 40.00 & 13.73 \\
    Positive\_regulation & 17.18 & 42.74 & 24.51 \\
    Negative\_regulation & 7.14 & 40.00 & 12.12 \\
    \hline
    Regulation Total & 13.65 & 42.14 & 20.62 \\
    \hline
    All Total & 26.77 & 56.22 & 36.27 \\
  \end{tabular}
  \end{small}
    \caption{Performance of the statistical model with $L_2$-regularized LR, using only features that can be converted to rules.}
  \label{tbl:res_lrl2}
  \vspace{-4mm}
\end{table}

For a fair comparison, we next trained the same model but using only features that can be converted to rules. As discussed, the features that were removed include bag-of-word features and features that count occurrences of tokens or entities in text. These results, summarized in Table~\ref{tbl:res_lrl2}, show that the overall F1 score drops 4 points. This suggests that rule languages need to be extended if they are to have the same representational power as feature-based models. Given that the focus of this work is not on the design of rule-based languages for IE, we will use this latter model as the starting point of our approach, ignoring (for now) the performance penalty observed above.

\begin{table}[t]
  \centering
  \begin{small}
  \begin{tabular}{l|r|r|r}
    Event Class & Recall & Precision & F1 \\
    \hline
    Gene\_expression & 58.71 & 78.28 & 67.09 \\
    Transcription & 37.80 & 55.36 & 44.93 \\
    Protein\_catabolism & 61.90 & 86.67 & 72.22 \\
    Phosphorylation & 46.81 & 84.62 & 60.27 \\
    Localization & 56.60 & 88.24 & 68.97 \\
    Binding & 16.13 & 33.33 & 21.74 \\
    \hline
    Event Total & 42.75 & 66.60 & 52.08 \\
    \hline
    Regulation & 8.88 & 65.22 & 15.62 \\
    Positive\_regulation & 13.13 & 40.50 & 19.83 \\
    Negative\_regulation & 8.16 & 55.17 & 14.22 \\
    \hline
    Regulation Total & 11.41 & 44.44 & 18.15 \\
    \hline
    All Total & 25.54 & 59.35 & 35.72 \\
  \end{tabular}
  \end{small}
      \caption{Performance of the statistical model with $L_1$-regularized LR, using only features that can be converted to rules.}
  \label{tbl:res_lrl1}
  \vspace{-4mm}
\end{table}

Importantly, a system with more than 1 million features is not interpretable.
To address this, we trained the same system using $L_1$ regularization as a form of
feature selection. This reduced the number of features with non-zero weights by two orders of magnitude: from over 1 million 
to 10,926. The performance of this model is shown in
Table~\ref{tbl:res_lrl1}.  The results demonstrate that this drastic reduction in the number of useful features came
with a small performance cost, of less than 1 F1 point. 

Given this successful compression of the feature space, we next convert this $L_1$-regularized model to rules, using the approach discussed in Section~\ref{sec:convert}. The performance of the rule-based model (before expert intervention!) is summarized in Table~\ref{tbl:res_rules}. The table shows that the overall cost of ``snapping to grid'' the statistical model is approximately 3 F1 points, which come from a drop in recall. This happens because many feature weights associated with specific labels (such as specific event triggers) have low values (due to sparsity), and, after the discretization process, the model can no longer prioritize these labels over the {\tt Nil} class. Interestingly, the same process yielded a small increase in precision from 59\% to 62\%. 


\begin{table}[t]
  \centering
  \begin{small}
  \begin{tabular}{l|r|r|r}
    Event Class & Recall & Precision & F1 \\
    \hline
    Gene\_expression & 55.34 & 76.95 & 64.38 \\
    Transcription & 28.05 & 53.49 & 36.80 \\
    Protein\_catabolism & 57.14 & 85.71 & 68.57 \\
    Phosphorylation & 40.43 & 90.48 & 55.88 \\
    Localization & 45.28 & 88.89 & 60.00 \\
    Binding & 12.90 & 33.33 & 18.60 \\
    \hline
    Event Total & 38.04 & 67.18 & 48.58 \\
    \hline
    Regulation & 5.33 & 75.00 & 9.94 \\
    Positive\_regulation & 10.70 & 48.89 & 17.55 \\
    Negative\_regulation & 5.61 & 55.00 & 10.19 \\
    \hline
    Regulation Total & 8.76 & 51.50 & 14.97 \\
    \hline
    All Total & 21.97 & 62.98 & 32.57 \\
  \end{tabular}
  \end{small}
  \caption{Performance of the rule-based model before expert intervention.}
  \label{tbl:res_rules}
    \vspace{-4mm}
\end{table}

\begin{table*}[t]
  \begin{small}
  \begin{tabular}{p{8cm}|p{7cm}}
    {\bf Suggested Change} & {\bf Description} \\
    \hline

    \multicolumn{2}{c}{{\bf Generalization}} \\

    \hline

    Add \newline \verb#/conj_(and|or|nor)|dep|cc|nn|prep_of/{,2}# to the end of Theme
    paths. & This transformation adds an optional modifier dependency
    to capture event participants when they appear either as nominal
    heads or modifiers. For example, because of this transformation,
    the model handles both these phrases similarly: ``phosphorylation
    of MEK'' and ``phosphorylation of the MEK protein'' \\
    \hline

    Ensure that all syntactic paths end in \verb#appos?#.
    & This change handles optional apposition to increase rule
    coverage. For example, in the sentence ``we found that A20
    binds to a novel protein, ABIN'', the word ABIN is an appositive
    for the word protein, so ABIN can serve as an argument in the binding event.  \\
    \hline

    Replace all specific named entities with their label. & For example,
    in rules such as \verb#[word=phosphorylates] (?=MEK)# that reference a
    specific protein, this replaces the specific protein (MEK) with the label
    \verb#Protein#. This improves rule generalization and, at the same time, reduces the
    total number of rules. \\
    \hline

    Make the \verb#>nn# dependency optional in \newline \verb#Theme:Protein = >nsubjpass >nn#.
    & The output of this transformation is similar to the first suggested change, i.e., the same rule captures event participants when they appear either as nominal heads or modifiers. \\
    \hline

    \multicolumn{2}{c}{{\bf Robustness}} \\

    \hline

    Replace \verb#agent# with \verb#/^(agent|prep_by)$/#.  & This
    modification is designed to account for a common parsing error of
    passive sentences, where \verb#agent# dependencies are incorrectly parsed as  \verb#prep_by#. \\
    \hline
    Change \verb#ccomp# to \verb#/(c|x)comp/# and \verb#acomp# to \verb#/(a|x)comp/#. & Parsers often confuse clausal and adjectival
    complements with open clausal complements. This transformation allows the rules to be robust to these errors. \\
    \hline

    \multicolumn{2}{c}{{\bf Readability}} \\

    \hline

    Merge rules when possible, e.g. \newline \verb#prep_of#, \verb#prep_of nn#,
    \verb#prep_of appos# \newline become
    \verb#prep_of (nn? appos | nn appos? nn?)?#. & This transformation collapses rules to
    improve readability. \\
    \hline

    Eliminate trigger rules that are not sufficiently discriminative
    (e.g., \verb#(?<=[lemma="be"]) [tag=/^(V|N|J)/#). & Some
    uninformative rules survived feature regularization but should be
    removed, as with the example rule which looks for any verb, noun, or adjective
    preceded by any conjugation of the verb ``be''. These rules
    inflate the grammar without adding discriminative power. \\
    \hline
    Do not use {\tt word} constraints. Only use {\tt lemma} and {\tt tag} features in trigger rules for simple events (other than transcription and
    binding). & This modification prefers lexical constraints on lemmas, because they generalize better than constraints on actual words. \\
    \hline

    Remove redundant constraints. & For example, in patterns like
    \verb#[incoming=nsubj & tag=/^N/]# the POS tag is redundant
    because it is implicitly defined through the incoming dependency
    (nominal subject). \\




  \end{tabular}
  \end{small}
  \caption{Representative examples of the rule changes suggested by
    linguistic experts.  }
  \label{tbl:changes}
    \vspace{-4mm}
\end{table*}

All in all, we consider a drop of 3 F1 points for the gain of interpretability an acceptable tradeoff. To empirically demonstrate  the value of interpretability, we let two Linguistics PhD students edit the generated rule-based model for one hour, aiming to improve its generalization, robustness to syntactic errors, and readability. 
The students were familiar with the Odin language~\cite{valenzuela2015} so they could ``read'' the model, and had a high-level understanding of the BioNLP shared task (although they did not participate in it). 
To guarantee that their recommendations came from understanding the model rather than other external factors, they were not given access to the BioNLP dataset. Given the large number of rules at this point, the students tended to randomly sample the rules in the model attempting to find repeated mistakes, rather than linearly inspect the list of rules. 
Table~\ref{tbl:changes} summarizes the experts' recommendations.  
As shown, several of the experts' suggestions involved
removing or collapsing rules, which reduced the number of rules from
10,926 to 8,868.  

Table~\ref{tbl:res_rules_human} lists the performance of the resulting model, after implementing the experts' recommendations.
The table shows that most of the F1 loss has been recovered: the overall F1 score for this system approaches 35 F1 points, and is less than 1 F1 point behind the $L1$-regularized LR statistical model. In addition of reducing the number of rules in the model, the experts' recommendations increased recall by over 4\%, which is more than what was lost during the conversion to rules. However, the precision of this configuration decreased by 11\%, which we blame on the experts' limited familiarity with the BioNLP task, and the strict settings of the experiment (no access to data, limited time). 
However, all in all, this experiment demonstrates that the rule-based model produced by the proposed approach is interpretable: the experts understood the model, and were able to improve it, both with respect to its generalization power and its readability. 

\begin{table}[t]
  \centering
  \begin{small}
  \begin{tabular}{l|r|r|r}
    Event Class & Recall & Precision & F1 \\
    \hline
    Gene\_expression & 60.39 & 70.49 & 65.05 \\
    Transcription & 31.71 & 57.78 & 40.94 \\
    Protein\_catabolism & 61.90 & 81.25 & 70.27 \\
    Phosphorylation & 42.55 & 86.96 & 57.14 \\
    Localization & 45.28 & 88.89 & 60.00 \\
    Binding & 22.18 & 23.50 & 22.82 \\
    \hline
    Event Total & 43.74 & 54.31 & 48.46 \\
    \hline
    Regulation & 10.06 & 40.48 & 16.11 \\
    Positive\_regulation & 12.80 & 44.89 & 19.92 \\
    Negative\_regulation & 10.71 & 51.22 & 17.72 \\
    \hline
    Regulation Total & 11.91 & 45.17 & 18.86 \\
    \hline
    All Total & 26.27 & 51.71 & 34.84 \\
  \end{tabular}
  \end{small}
  \caption{Performance of the rule-based model, after expert intervention.}
  \label{tbl:res_rules_human}
\end{table}

\begin{figure}[t]
  \centering
  \includegraphics[scale=0.35]{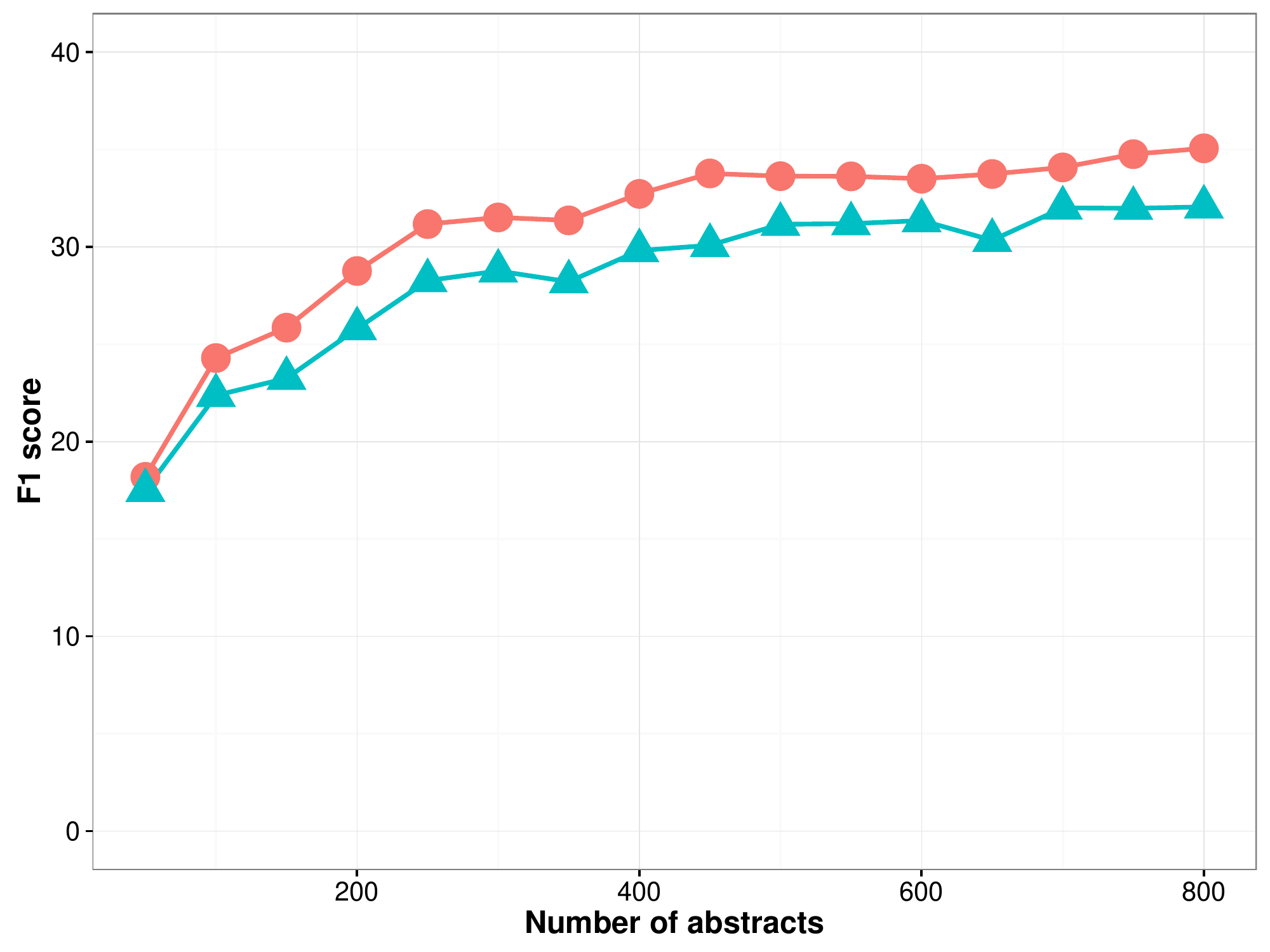}
  \caption{Learning curve showing the change in F1 performance as a
    function of the amount of training data.  We compare the performance of
    the $L_1$-regularized logistic regression (shown using circles) with
    the rule-based model prior to the expert intervention (shown using triangles).}
  \label{fig:learningcurve}
    \vspace{-4mm}
\end{figure}

Lastly, Figure~\ref{fig:learningcurve} shows a learning curve for the statistical model and the corresponding rule-based model (before expert intervention). 
The curve shows that the rule-based model follows closely the behavior of its statistical counterpart, with a small penalty of 1-2 F1 points throughout. As discussed before, this performance loss can be mitigated through interventions by domain experts. 

\section{Related Work}

Most of the biomedical IE systems in academia rely on supervised machine learning.
This includes the top performing system at the BioNLP 2009 shared task~\cite{bjorne2009extracting}, as well as several following approaches that improve upon its performance~\cite{miwa2010event,McClosky:12,miwa2012boosting,bui2012robust,venugopal2014relieving}.

However, rule-based approaches~\cite{Appelt:93,cunningham2002framework,piskorskishallow,li2011systemt,tokensregex2014}  are preferable when
the corresponding systems have to be deployed for long periods of time, during which
they have to be maintained and improved. 
This has been recognized in industry~\cite{chiticariu2013}. 

We bring together these two diverging directions by combining the advantages of
ML with the interpretability of rule-based
approaches. By representing the model as a collection of declarative rules, experts
can directly edit the model, thus guaranteeing that the desired changes are
actually applied. This is in contrast with methods such as active
learning, in which the learning algorithm presents the ``human in the loop'' with new
examples to annotate~\cite{thompson1999active}. 
Although active learning may require less domain expertise than our proposal, 
it generally does not guarantee that the examples provided are actually propagated in the model (the learning algorithm may choose to override them with other data).

\section{Conclusion and Future Work}

We have proposed a simple approach that marries the advantages of machine learning models for information extraction (such as learning directly from data) with the benefits of rule-based approaches (interpretability, easier maintainability). Our approach starts by training a feature-based statistical model, then converts this model to a rule-based variant by converting its features to rules and its feature weights to discrete votes. In doing so, our proposal learns from data similar to other machine learning approaches, but produces an interpretable rule-based model that can be directly edited by experts. Using the BioNLP 2009 event extraction task as a test bed, we show that while there is a small performance penalty when converting the statistical model to rules, the gain in interpretability compensates for that. 

In this work, we focused on building upon feature-based classifiers, in particular logistic regression, 
due to their potential extensions to distant supervision (DS), where training data is generated automatically 
by aligning a knowledge base (KB) of known examples (e.g., known drug-gene interactions) with text (e.g., scientific publications).
Distant supervision has obvious applications to bioinformatics~\cite{craven1999constructing}, but it generally suffers from noise in the automatically-generated annotations~\cite{riedel2010modeling}. In future work, we plan to combine our work with distant supervision by adapting our proposal to logistic regression variants that are robust to the noise introduced in DS~\cite{surdeanu2012multi}. This extension would make it possible
to generate rules even when no annotated examples are available, as
long as a suitable KB of known examples exists.

Another planned extension of this work focuses on reducing the
number of generated rules by merging/collapsing similar paths into a single pattern.
This can be achieved by constructing a minimal
deterministic acyclic finite-state automaton
(DAFSA)~\cite{daciuk2000incremental} with the paths that are similar,
and then converting the DAFSA into a single
pattern~\cite{neumann2005converting}. For example, such approaches would collapse the two patterns: {\tt dobj} and {\tt dobj nn}, into a single one: {\tt dobj nn?}. This is fundamental for the long-term maintainability of the rule-based model, because the human experts would have to maintain considerably fewer rules.

Lastly, we plan to improve the ``snap to grid'' algorithm.
Currently, the conversion of weights to votes is implemented
using Scott's rule~\cite{scott1979optimal}, which is one method among
several available to choose a histogram's bin size. Scott's method assumes that
all bins have the same size, which may not be the best solution if interpretability is the goal. A
potentially better approach is to select the bin divisions in a
way that retains as much of the information contained in the weights
as possible, while minimizing the number of bins.

\section*{Acknowledgments}

This work was funded by the Defense Advanced Research Projects Agency (DARPA) Big Mechanism program under ARO contract W911NF-14-1-0395.


\bibliography{acl2016}
\bibliographystyle{acl2016}

\end{document}